%% file: main.tex
\documentclass[10pt,twocolumn,letterpaper]{article}
\usepackage[pagenumbers]{cvpr}
\usepackage{times}
\usepackage{latexsym}
\usepackage{graphicx}
\usepackage{tabularx}
\usepackage{xspace}
\usepackage{xcolor}
\usepackage[table,xcdraw]{xcolor}
\usepackage{tcolorbox}
\definecolor{cvprblue}{rgb}{0.21,0.49,0.74}
\definecolor{my_green}{RGB}{46, 151, 78}
\definecolor{my_red}{RGB}{216, 37, 34}
\usepackage[pagebackref,breaklinks,colorlinks,allcolors=cvprblue]{hyperref}
\usepackage{makecell}
\newcommand{\NAME}{\textsc{MIRA}\xspace}
\newcommand{\DATA}{\textsc{MIRA-Editing}\xspace}
\title{\NAME: Multimodal Iterative Reasoning Agent for Image Editing}

\author{Ziyun Zeng\textsuperscript{1 }, Hang Hua\textsuperscript{2 }, Jiebo Luo\textsuperscript{1 }\\
\textsuperscript{1}University of Rochester, \textsuperscript{2}MIT-IBM Watson AI Lab\\
{\tt\small ziyun.zeng@rochester.edu, hang.hua1@ibm.com, jluo@cs.rochester.edu}
}

\begin{document}
\maketitle
\input{sec/0_abstract}    
\input{sec/1_intro}

\input{sec/2_relatedwork}
\input{sec/3_dataset}
\input{sec/4_method}

\input{sec/5_experiment}
\input{sec/6_conclusion}
{
    \small
    \bibliographystyle{ieeenat_fullname}
    \bibliography{main}
}

\end{document}

%% file: sec/0_abstract.tex
\begin{abstract}
Instruction-guided image editing offers an intuitive way for users to edit images with natural language. However, diffusion-based editing models often struggle to accurately interpret complex user instructions—especially those involving compositional relationships, contextual cues, or referring expressions—leading to edits that drift semantically or fail to reflect the intended changes. We tackle this problem by proposing \textbf{\NAME} (\textbf{M}ultimodal \textbf{I}terative \textbf{R}easoning \textbf{A}gent), a lightweight, plug-and-play multimodal reasoning agent that performs editing through an iterative perception–reasoning–action loop, effectively simulating multi-turn human–model interaction processes. Instead of issuing a single prompt or static plan, \NAME predicts atomic edit instructions step by step, using visual feedback to make its decisions. Our 150K multimodal tool-use dataset, \DATA, combined with a two-stage SFT + GRPO training pipeline, enables \NAME to perform reasoning and editing over complex editing instructions. When paired with open-source image editing models such as Flux.1-Kontext, Step1X-Edit, and Qwen-Image-Edit, \NAME significantly improves both semantic consistency and perceptual quality, achieving performance comparable to or exceeding proprietary systems such as GPT-Image and Nano-Banana. 
\end{abstract}

%% file: sec/1_intro.tex
\section{Introduction}
\label{sec:intro}

As visual generation systems become increasingly interactive, instruction-guided image editing has attracted growing attention for its ability to follow user instructions to perform precise, controllable visual manipulations. Recent advanced diffusion-based editing models, such as Qwen-Image-Edit~\cite{wu2025qwen}, Flux.1-Kontext~\cite{batifol2025flux}, and Step1X-Edit~\cite{liu2025step1x}, have demonstrated promising progress in aligning textual guidance with pixel-level transformations. However, these systems exhibit significant degradation when handling complex instructions that involve compositional reasoning, multiple-object interactions, or context-dependent manipulations. Even proprietary multimodal systems like Seedream 4.0~\cite{seedream2025seedream}, GPT-Image~\cite{openai2025gpt_image}, or Nano-Banana~\cite{google2025nano_banana}, though more capable, still struggle with maintaining fine-grained consistency and controllability in complex instruction-guided image editing. These limitations highlight the need for editing systems that can go beyond one-shot prompt execution and instead reason, adapt, and revise their edits interactively.

To mitigate the discrepancy between user intent and editing outcomes, recent research has moved beyond purely text-to-edit systems and explored embedding reasoning and orchestration via vision–language models. One line of research augments editing pipelines by enriching text embeddings or refining user instructions before forwarding them to diffusion-based editors~\cite{cai2025hidream,fu2023guiding,xia2024llmga,yu2024promptfix,zhang2025context,zhao2024ultraedit,zhou2025fireedit}. These approaches improve textual alignment and enhance editing controllability, but they remain constrained to static prompt refinement.
A complementary and more recent direction reframes instruction-guided image editing as a reasoning-driven orchestration problem. Instead of executing a single prompt once, a vision-language model acts as an editing agent that decomposes complex image editing instructions into semantically coherent sub-tasks, plans an execution sequence, and coordinates a suite of specialized expert models such as localization, inpainting, editing, composition, and global transformation. These agentic frameworks, characterized by programmatic decomposition, iterative reasoning, and multi-tool collaboration, achieve notable improvements on compositional, multi-step, and context-dependent editing tasks by integrating structured reasoning with iterative visual feedback mechanisms~\cite{gupta2025costa,hu2025image,ji2025instruction,hua2024finematch,liang2025llm,wang2024genartist,yeh2025beyond}.
Although these architectures exhibit strong compositional reasoning and interpretability, they typically rely on large toolchains and complex model coordination, making them computationally intensive and difficult to scale within open-source ecosystems.

\begin{figure*}[th]
  \centering
   \includegraphics[width=0.98\textwidth]{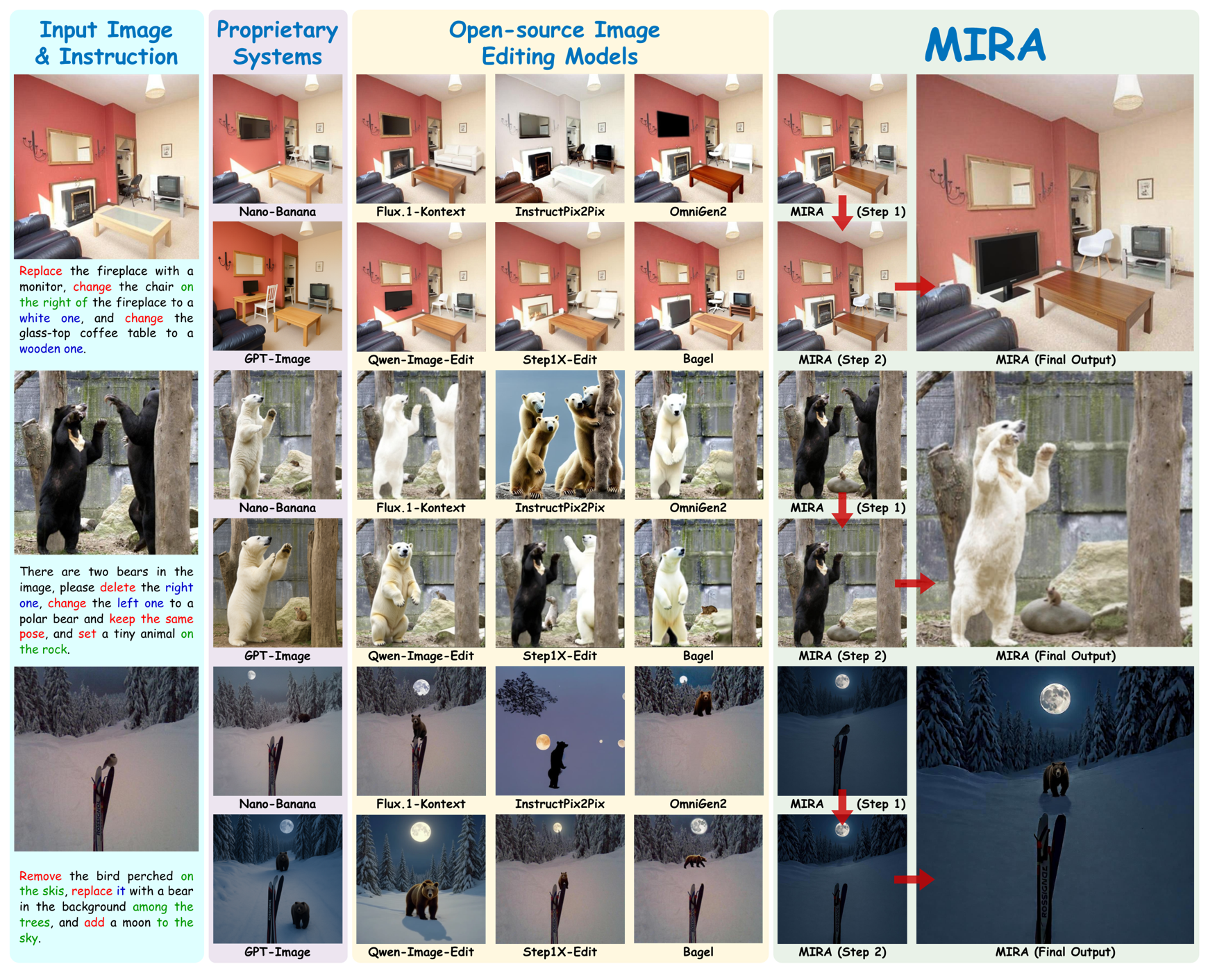}
    \vspace{-2mm}
   \caption{Qualitative comparison of \NAME against leading proprietary and open-source image editing models on complex instructions. The rightmost column illustrates \NAME's unique iterative reasoning and editing process, displaying the intermediate visual results after each step of its perception-reasoning-action loop.}
   \label{fig:qual}
   \vspace{-6mm}
\end{figure*}

Building upon these insights, we introduce \NAME, a lightweight, plug-and-play vision–language agent that formulates instruction-guided image editing as an iterative visual reasoning process rather than a static orchestration pipeline. Unlike prior editing systems that pre-plan an entire sequence of operations or rely on handcrafted tool chains, \NAME performs step-wise agentic action selection: at each iteration, the vision-language model (VLM) observes the original image, the user instruction, and the intermediate editing state, reasons about the remaining visual discrepancy, and predicts the next atomic editing action or a stop decision. This cycle defines a simple yet effective agentic loop: state $\rightarrow$ multimodal reasoning $\rightarrow$ action $\rightarrow$ environment feedback,
where the “environment” corresponds to the visual editor that executes the predicted action. In this loop, reasoning is grounded in editing results; the agent learns and adapts by observing the discrepancy between the instruction and editing results. This cycle repeats iteratively until the complex instruction is fully satisfied.

Through iterative reasoning and feedback-driven tool use, \NAME effectively improves editing performance in semantic consistency and perceptual quality through iterative multimodal reasoning and feedback-driven tool calling. By casting image editing as an agentic loop, \NAME enables open-source diffusion-based editors to handle complex, ambiguous, and context-dependent instructions with both interpretability and efficiency. This design significantly boosts the capabilities of open-source editing models, often allowing them to match or even surpass the performance of proprietary systems.

Our main contributions are summarized as follows:
\begin{itemize}
    \item A lightweight, agentic, and plug-and-play vision–language model that can be seamlessly paired with existing open-source image editing backbones. By combining multimodal reasoning with explicit tool-use, \NAME empowers these models to handle complex editing instructions. It significantly reduces the performance gap between open-source and proprietary image editing systems.
    \item A large-scale training dataset, \DATA, tailored for training automatic tool-use models for instruction-guided image editing. The dataset comprises 150,000 high-quality paired examples generated through hierarchical instruction aggregation, semantic rewriting, and ranking-based filtering, providing diverse and reliable multimodal supervision.
    \item A two-stage training pipeline combining SFT and GRPO. During the GRPO phase, we incorporate a novel composite reward model that couples an image-editing backbone with an image-editing reward model to assess the quality and fidelity of edit instructions. This design provides richer, more reliable, and semantically grounded optimization signals.
\end{itemize}

%% file: sec/2_relatedwork.tex
\section{Related Works}
\label{sec:related}

\subsection{Instruction-guided Image Editing}
Instruction-guided image editing enables precise and intuitive manipulation of visual content through natural language instructions. 
Early prompt-based diffusion editing methods~\cite{couairon2022diffedit, hertz2022prompt,kawar2023imagic,meng2021sdedit} demonstrated the potential of prompt-level control for semantic image manipulation by leveraging textual guidance to achieve localized and structurally coherent edits without additional training. These approaches revealed the expressive power of diffusion models in responding to prompt semantics, yet remained limited by their reliance on manually crafted text descriptions. 
Building on this foundation, InstructPix2Pix~\cite{brooks2023instructpix2pix} proposed a data-driven approach that learns direct text-to-edit mappings, inspiring a wave of instruction-based editing methods such as MagicBrush~\cite{zhang2023magicbrush}, UltraEdit~\cite{zhao2024ultraedit}, InstructDiffusion~\cite{geng2024instructdiffusion}, InstructCV~\cite{gan2023instructcv}, and PromptFix~\cite{yu2024promptfix}, which emphasize large-scale instruction–image alignment, as well as precise object-level manipulation and compositing~\cite{song2023objectstitch,song2024imprint,song2024refine,yu2025omnipaint,zhengpixperfect}.
To strengthen instruction understanding and contextual reasoning, advanced systems such as MGIE~\cite{fu2023guiding}, SmartEdit~\cite{huang2024smartedit}, InstructEdit~\cite{wang2023instructedit}, and OmniGen2~\cite{wu2025omnigen2} leverage large language or vision–language models to enhance semantic interpretation, context awareness, and fine-grained control in instruction-guided image editing.
Meanwhile, open-source models such as Qwen-Image-Edit~\cite{wu2025qwen}, Flux.1-Kontext~\cite{batifol2025flux}, and Step1X-Edit~\cite{liu2025step1x} have advanced diffusion-based editing capabilities, combining high-resolution latent diffusion frameworks with robust controllability. In contrast, proprietary systems including Seedream 4.0~\cite{seedream2025seedream}, GPT-Image~\cite{openai2025gpt_image}, and Nano-Banana~\cite{google2025nano_banana} represent the state of the art in multimodal reasoning, demonstrating superior fidelity, responsiveness, and adaptability in instruction-guided image editing.

\subsection{MLLMs for Image Editing}
Multimodal large language models (MLLMs) \cite{guo2025deepseek,hua2025v2xum,hua2025finecaption,hu2023promptcap,hua2024mmcomposition,hua2025mmig,meta2025llama,sun2025latent,tang2025video,yu2024chain,zeng2025glm,zhu2025internvl3} have recently emerged as powerful engines for reasoning-driven image generation and editing. Early studies leverage MLLMs primarily for text-to-image synthesis, where MLLMs interpret natural language descriptions and generate creative visual content in a zero-shot manner~\cite{chen2025t2i,huang2024dialoggen,wang2024genartist}. These approaches highlight the strong cross-modal understanding of MLLMs but remain limited to one-shot generation without explicit control over iterative editing.
To support instruction-guided editing, several studies employ MLLM prompt optimizers that rewrite or enrich user instructions before passing them to diffusion-based image editing models, such as MGIE~\cite{fu2023guiding}, LLMGA~\cite{xia2024llmga}, PromptFix~\cite{yu2024promptfix}, and HiDream-E1~\cite{cai2025hidream}. This strategy enhances linguistic clarity and improves alignment between textual descriptions and visual execution. However, such methods operate in a static manner. Once a refined prompt is produced, the model does not evaluate whether the edited result aligns with the intended instruction, limiting its adaptability to complex or ambiguous instructions.
More recent works extend MLLMs into agentic frameworks that decompose editing tasks and coordinate external tools. While agentic reasoning has shown remarkable success in scientific domains~\cite{cao2024presto,lin2024battleagent,tang2025cellforge,tang2025chemagent,tang2025medagentsbench,zeng2025use,zeng2026automated}, applying it to visual editing remains challenging. Systems such as X-Planner~\cite{yeh2025beyond}, RefineEdit-Agent~\cite{liang2025llm}, and CoSTA*~\cite{gupta2025costa}. These systems introduce structured reasoning and sequential planning for complex editing workflows.

%% file: sec/3_dataset.tex
\section{Data Curation}
\label{sec:dataset}

\begin{figure*}[th]
  \centering
   \includegraphics[width=\textwidth]{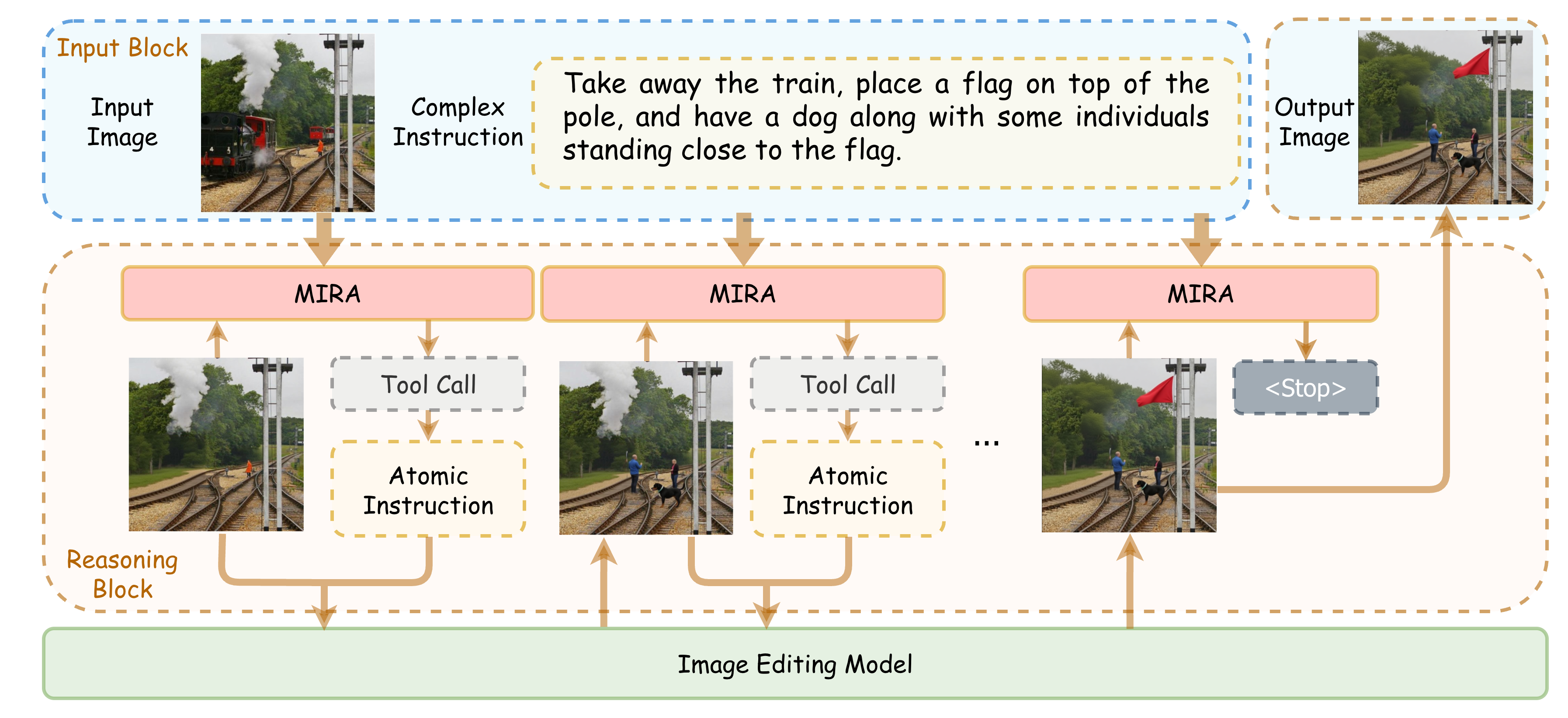}

   \caption{\textbf{Workflow of our multimodal reasoning and editing agent \NAME.} Given an input image and a complex natural-language instruction, \NAME engages in an iterative perception–reasoning–action loop. At each step, the agent analyzes the current visual state and textual context to generate an atomic edit instruction, which is executed by an external image-editing model. The updated image is fed back into the agent to guide the next step. This loop continues until the full instruction is satisfied, yielding the final edited result.}
   \label{fig:agent_workflow}
   \vspace{-4mm}
\end{figure*}

High-quality and compositionally rich data is essential for training \NAME. Due to the lack of publicly available data that satisfies the need for multi-step reasoning supervision, complex instruction alignment, and high-fidelity image editing trajectories, we construct a new dataset with 150K paired samples. We also provide detailed quantitative and qualitative analyses in the supplementary material.
\subsection{Data Preprocessing Pipeline}

\noindent \textbf{Instruction Aggregation.}
We transform the multi-turn editing sequences in SeedEdit~\cite{ge2024seed} into single complex instructions by merging their atomic edits. To further enrich compositional complexity, we generate in-order and permuted variants, allowing the model to learn from a broader range of compositional structures.

\noindent \textbf{Two-Level Instruction Rewriting.}
To enhance linguistic diversity, we paraphrase instructions at two levels: atomic edits are paraphrased individually, and the full aggregated instruction is holistically rewritten with varied structures and connectives, yielding $x$ semantically equivalent yet stylistically diverse variants.

\noindent \textbf{Candidate Generation.}
Each rewritten instruction is executed using a strong open-source image editing model, producing $x$ candidate edited trajectories. These variations expose differences caused by paraphrasing and provide a diverse pool of potential supervision signals.

\noindent \textbf{Semantic Consistency Ranking.}
To filter noisy or misaligned outputs, we evaluate all $x$ candidates using VieScore~\cite{ku2023viescore}, choosing Gemini-2.5-Flash and Qwen2.5-VL-72B-Instruct as backbone. They rank candidates by semantic consistency with the original complex instruction, and we retain only the top-1 result.

\noindent \textbf{Final Sample Construction.}
The final dataset entry consists of the input image, the selected high-quality edited trajectory, the aggregated complex instruction, and its rewritten variants. This ensures each sample provides clean, diverse, and semantically faithful supervision for iterative reasoning and editing.

\subsection{Training Data Formulation.}
To train \NAME as an iterative reasoning agent in a perception-reasoning-action loop, each curated editing trajectory is decomposed into step-wise supervision. Given a complex instruction $C$, the original image $I_0$, and the intermediate result $I_{t-1}$, the target atomic edit $u_t$ and its execution via image editing $I_t$ form a transition:
\begin{equation}
    (I_{t-1}, I_0, C) \rightarrow u_t \rightarrow I_t
\end{equation}
We convert each trajectory into three structured data types. \textbf{Type~1 (Start)} samples supervise the first-step prediction conditioned only on $I_0$ and $C$. \textbf{Type~2 (Continue)} samples, which constitute the majority of the dataset, train the model to iteratively refine edits based on visual feedback from $I_{t-1}$. \textbf{Type~3 (Stop)} samples supervise termination detector during perception, enabling the agent to recognize task completion and avoid unnecessary edits. This structured formulation shows \NAME's inference time editing loop and provides compact and effective multimodal supervision for progressive, feedback-driven reasoning.

\begin{figure}[h]
  \centering
   \includegraphics[width=\linewidth]{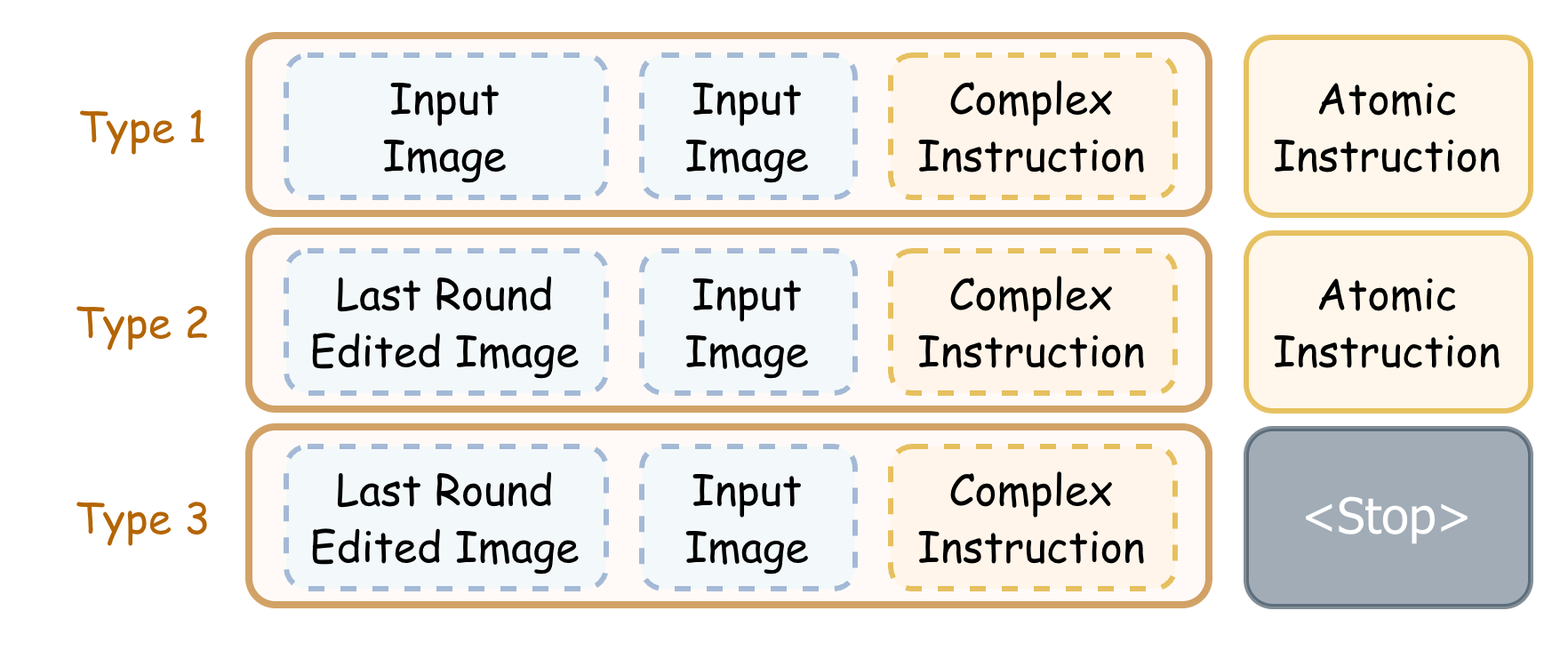}

   \caption{Three types of editing samples in \DATA.}
   \label{fig:dataset_formulation}
   \vspace{-6mm}
\end{figure}

%% file: sec/4_method.tex
\section{Methodology}
\label{sec:method}

\subsection{The MIRA Framework}
\begin{figure*}[th]
  \centering
   \includegraphics[width=0.95\textwidth]{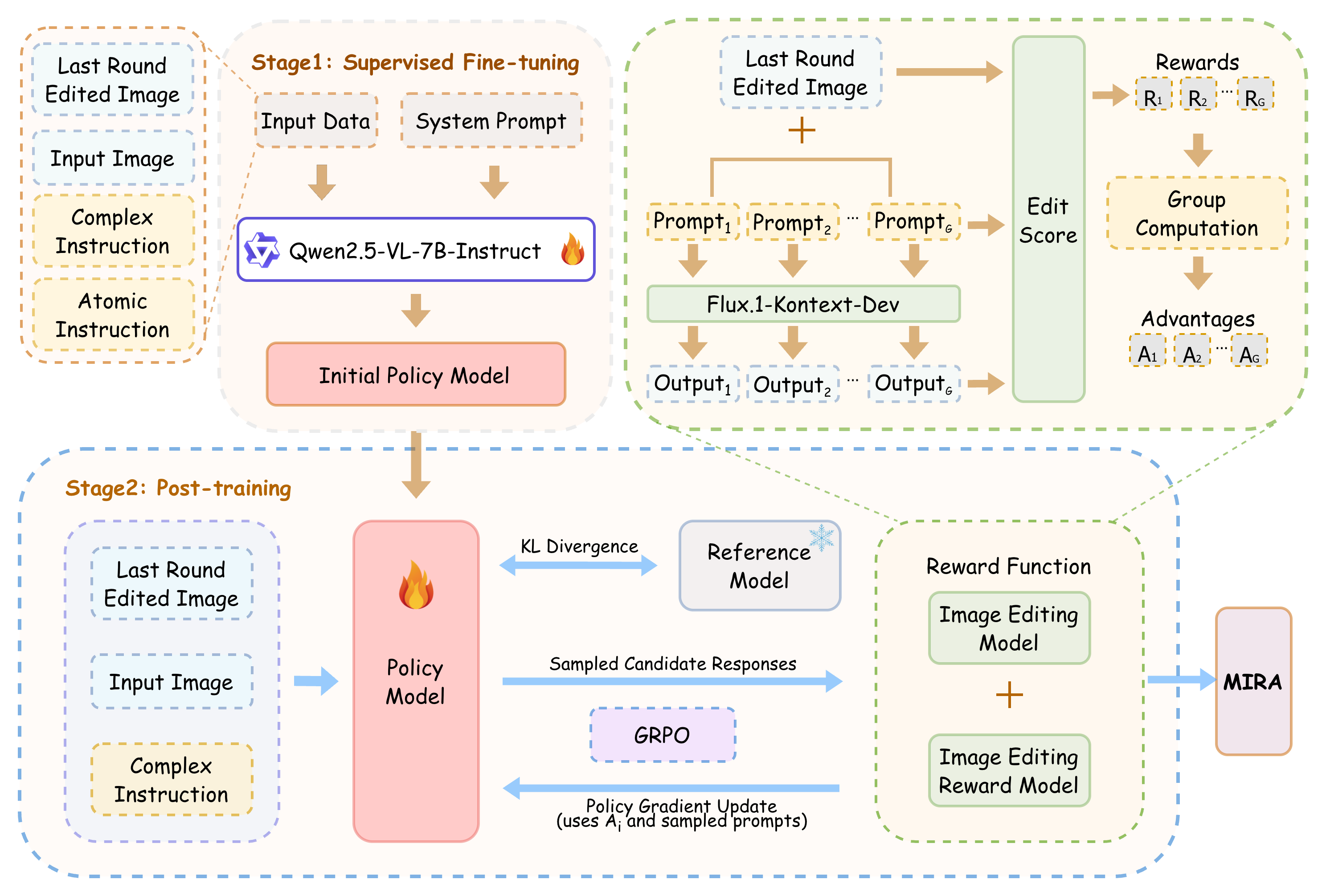}

   \caption{\textbf{Overview of the \NAME Training Pipeline.} The training pipeline comprises two stages: (1) Supervised Fine-Tuning and (2) Reinforcement Learning. Stage 1 fine-tunes Qwen2.5-VL-7B-Instruct on paired samples of the input image, the previously edited image, and the complex instruction to initialize the policy model. Stage 2 applies GRPO to further refine the policy, using a composite reward function that couples an image editing model with an editing reward model to score edit quality and provide optimization signals.}
   \label{fig:training_pipeline}
   \vspace{-4mm}
\end{figure*}
Our proposed \NAME, as shown in Figure \ref{fig:agent_workflow}, builds upon Qwen2.5-VL~\cite{bai2025qwen2}, leveraging its strong multimodal reasoning and visual understanding capabilities. The model is trained through supervised fine-tuning (SFT) followed by reinforcement learning post-training to improve semantic consistency and perceptual quality through reward-driven optimization.
At inference time, \NAME performs instruction-guided editing in a strictly iterative and closed-loop manner, issuing exactly one atomic editing instruction at each step and dynamically adapting its plan based on the most recent visual feedback.

Formally, given an input image $I_0$ and a complex instruction $C$, the agent maintains the current intermediate result $I_{t-1}$ and, at step $t$, predicts the next editing instruction as
\begin{equation}
u_t = \pi_\theta(I_{t-1}, I_0, C)
\label{eq:policy}
\end{equation}
where $\pi_\theta$ denotes the policy induced by Qwen2.5-VL after SFT and RL post-training. 
The predicted instruction $u_t$ is then executed by a diffusion-based image editing model $E_t \in \mathcal{E}$:
\begin{equation}
I_t = E_t(I_{t-1}, u_t)
\end{equation}
where $\mathcal{E}$ represents a family of interchangeable open-source editors such as Flux.1-Kontext-Dev~\cite{batifol2025flux}, Qwen-Image-Edit~\cite{wu2025qwen}, and Step1X-Edit~\cite{liu2025step1x}. 
After each step, a lightweight termination controller determines whether the process should continue:
\begin{equation}
d_t = \tau_\phi(I_t, I_0, C) \in \{\texttt{continue}, \texttt{stop}\}
\end{equation}
where $\tau_\phi$ is trained jointly with the policy to predict task completion. If $d_t = \texttt{stop}$, the procedure halts and returns $I_t$ as the final output. Otherwise, the triplet $(I_t, I_0, C)$ is fed back into the policy to produce the next instruction $u_{t+1}$.

In contrast to prior agentic systems that predict the entire editing sequence in one shot, \NAME performs adaptive, feedback-driven instruction planning and goal-conditioned refinement. This receding-horizon approach naturally yields a sequence of semantically atomic operations ${u_t}$, each delegated to an interchangeable editing backend. The design integrates the reasoning strengths of vision–language models with the modularity of open-source diffusion-based editors, enabling scalable and controllable execution of complex editing instructions without extensive multi-tool orchestration.

\begin{table*}[t]
    \centering
    \caption{Quantitative comparison of baseline vs. \NAME-enhanced image editing models in terms of semantic consistency and perceptual quality. GPT-SC, Gemini-SC, and Qwen3VL-SC report semantic consistency measured by GPT-5, Gemini-2.5-Flash, and Qwen3-VL-30B-A3B-Instruct using the viescore prompt. ARNIQA and TOPIQ assess perceptual quality via human preference modeling and traditional IQA metrics. EditScore-SC, EditScore-PQ, and EditScore-OA provide open-source semantic consistency, perceptual quality, and overall alignment scores. The \textbf{best} and \underline{second best} results are in bold and underlined, respectively, excluding proprietary systems.}
    \label{tab:quan_compare_insert}
    \vspace{-3mm}
    \resizebox{\textwidth}{!}{
    \begin{tabular}{l|cccccccc}
        \toprule
        \textbf{Method} & \textbf{GPT-SC} $\uparrow$ & \textbf{Gemini-SC} $\uparrow$ & \textbf{Qwen3VL-SC} $\uparrow$ & \textbf{EditScore-SC} $\uparrow$& \textbf{ARNIQA} $\uparrow$ & \textbf{TOPIQ} $\uparrow$  & \textbf{EditScore-PQ} $\uparrow$  & \textbf{EditScore-OA} $\uparrow$ \\
        \cmidrule(r){1-1} \cmidrule(r){2-5} \cmidrule(r){6-8} \cmidrule(r){9-9}
        \rowcolor[HTML]{e9edf6}
        & \multicolumn{8}{c}{\textit{Open Source Image Editing Models}} \\
        \midrule
        InstructPix2Pix~\cite{brooks2023instructpix2pix} & 1.804 & 2.298 & 2.166 & 6.414 & 0.511 & 0.277 & 4.539 & 4.937\\
        OmniGen2~\cite{wu2025omnigen2} & 4.750 & 5.204 & 4.760 & 6.867 & 0.525 & 0.310 & 4.913 & 5.468\\
        Bagel~\cite{deng2025emerging} & 6.228 & 6.482 & 5.754 & 7.581 & 0.507 & 0.283 & 5.244 & 5.949\\
        Step1X-Edit~\cite{liu2025step1x} & 5.942 & 6.440 & 4.946 & 8.542 & 0.507 & 0.278 & 5.474 & 6.636\\
        Flux.1-Kontext~\cite{batifol2025flux} & 5.908 & 6.290 & 5.116 & 8.194 & 0.563 & 0.330 & 5.673 & 6.507\\
        Qwen-Image-Edit~\cite{wu2025qwen} & 6.290 & \underline{6.934} & 5.792 & 7.879 & 0.582 & 0.341 & 
        \textbf{6.119} & 6.625\\
        \midrule
        \rowcolor[HTML]{e9edf6}
        & \multicolumn{8}{c}{\textit{Proprietary Image Editing Models}} \\
        \midrule
        GPT-Image~\cite{openai2025gpt_image} & 7.410 & 7.232 & 6.656 & 8.629 & 0.635 & 0.383 & 7.666 & 7.977\\
        Nano-Banana~\cite{google2025nano_banana} & 7.972 & 8.124 & 6.514 & 8.924 & 0.599 & 0.318 & 6.740 & 7.551\\
        \midrule
        \rowcolor[HTML]{e9edf6}
        & \multicolumn{8}{c}{\textit{Open Source Image Editing Models Plug and Play with \NAME }} \\
        \midrule
        \rowcolor{violet!6}
        \textbf{Step1X-Edit + \NAME}  & \underline{6.322} & 6.886 & \underline{5.844} & \textbf{8.668} & 0.601 & 0.296 & 5.640 & \underline{6.800}\\
        \rowcolor{violet!10}
        \textbf{Flux.1-Kontext + \NAME}  & 6.202 & 6.670 & 5.802 & 8.532 & \textbf{0.619} & \underline{0.353} & 5.473 & 6.610\\
        \rowcolor{violet!14}
        \textbf{Qwen-Image-Edit + \NAME} & \textbf{6.882} & \textbf{7.104} & \textbf{6.280} & \underline{8.583} & \underline{0.612} & \textbf{0.357} & \underline{5.801} & \textbf{6.871}\\
        \bottomrule
    \end{tabular}
    }
    \vspace{-4mm}
\end{table*}

\subsection{Training Pipeline}

As shown in Figure~\ref{fig:training_pipeline}, \NAME is trained via a two-stage pipeline that gradually shifts from imitation-based instruction following to reward-driven multimodal reasoning. In Stage~1, we initialize the policy model $\pi_{\theta}$ from Qwen2.5-VL-7B-Instruct and train it on curated instruction–image triplets to generate atomic editing actions as~\eqref{eq:policy}.
This stage focuses on imitation learning from high-quality paired data, aligning visual transformations with textual intent.

In Stage~2, we apply Group Relative Policy Optimization (GRPO)~\cite{shao2024deepseekmath} in a step-wise manner. At each step $t$, the policy model generates multiple atomic edit instructions $\{u_t^k\}_{k=1}^{K}$ conditioned on the current visual state $I_{t-1}$. Each atomic edit instruction is executed by the external image model, such as Flux.1-Kontext, to produce an updated image $I_t^k = E(I_{t-1}, u_t^k)$,  and the quality of the edit is assessed by a fixed reward model, such as EditScore.
$r_t^k = R(I_t^k, I_{t-1}, u_t^k)$ measures both semantic consistency and perceptual quality. Therefore, the reward can be formulated as 
\begin{equation}
r_t^k = \lambda_{\mathrm{sc}}\, r_{\mathrm{sc}}(I_t^k, I_{t-1}, u_t^k)
      + \lambda_{\mathrm{pq}}\, r_{\mathrm{pq}}(I_t^k, I_{t-1}, u_t^k),
\end{equation}
where $r_{\mathrm{sc}}$ and $r_{\mathrm{pq}}$ quantify semantic consistency and perceptual quality.
The resulting rewards $\{r_t^k\}$ are normalized into advantages $\{A_t^k\}$, and the policy is optimized based on all step-level transitions $(I_{t-1}, u_t^k, r_t^k)$. The update rule for the policy model is formulated as
\begin{equation*}
\begin{split}
&\nabla_{\theta}\mathcal{J}_t(\theta)
= \mathbb{E}_{k}\!\left[
    A_t^{k}\,\nabla_{\theta}
    \log\pi_{\theta}\!\big(u_t^k \mid I_{t-1}, r_t^k\big)
  \right] \\
& - \beta\,
  D_{\mathrm{KL}}\!\Big(
  \pi_{\theta}(\cdot \mid I_{t-1}, u_t^k, r_t^k)
  \,\big\|\, 
  \pi_{\text{ref}}(\cdot \mid I_{t-1}, u_t^k, r_t^k)
  \Big)
\end{split}
\end{equation*}

where $\pi_{\text{ref}}$ denotes the frozen, supervised fine-tuned policy model used as the reference model. Both $E$ and $R$ are used only during the forward evaluation and are not backpropagated.

%% file: sec/5_experiment.tex
\section{Experiment}
\label{sec:experiment}
\subsection{Experiment Settings}

\begin{table*}[th]
    \centering
    \caption{Quantitative comparison of \NAME-enhanced versus other VLM-enhanced image editing models on semantic consistency and perceptual quality. Metric definitions follow Table~\ref{tab:quan_compare_insert}. All models are evaluated in a plug-and-play setting with Flux.1-Kontext-Dev and Qwen-Image-Edit as the base instruction-guided editing backbones.}
    \label{tab:quan_compare_vlm}
    \vspace{-2mm}
    \resizebox{\linewidth}{!}{
    \begin{tabular}{l|cccccccc}
        \toprule
        \textbf{Method} & \textbf{GPT-SC} $\uparrow$ & \textbf{Gemini-SC} $\uparrow$ & \textbf{Qwen3VL-SC} $\uparrow$ & \textbf{EditScore-SC} $\uparrow$ & \textbf{ARNIQA} $\uparrow$ & \textbf{TOPIQ} $\uparrow$ & \textbf{EditScore-PQ} $\uparrow$  & \textbf{EditScore-OA} $\uparrow$ \\
        \cmidrule(r){1-1} \cmidrule(r){2-5} \cmidrule(r){6-8} \cmidrule(r){9-9}
        \rowcolor[HTML]{e9edf6}
        & \multicolumn{8}{c}{\textit{Plug and Play with Flux.1-Kontext-Dev}} \\
        \midrule
        Qwen3-VL-8B~\cite{bai2025qwen2}  & 5.488 & 5.688 & 4.264 & 8.243 & 0.616 & 0.317 & 5.226 & 6.279\\
        Qwen3-VL-30B~\cite{bai2025qwen2}  & 3.792 & 3.920 & 2.808 & 8.330 & 0.583 & 0.306 & \textbf{5.729} & \underline{6.455}\\
        GPT-5~\cite{openai2025gpt_4o}  & \textbf{7.074} & \underline{6.566} & \underline{5.610} & \underline{8.358} & \textbf{0.653} & \textbf{0.378} & 5.129 & 6.284\\
        \rowcolor{violet!6}
        \textbf{\NAME  3B} & 5.628 & 6.422 & 5.754 & 8.168 & 0.615 & 0.338 & 5.094 & 6.177\\
        \rowcolor{violet!10}
        \textbf{\NAME  7B} & \underline{6.202} & \textbf{6.670} & \textbf{5.802} & \textbf{8.532} & \underline{0.619} & \underline{0.345} & \underline{5.473} & \textbf{6.610}\\
        \midrule
        \rowcolor[HTML]{e9edf6}
        & \multicolumn{8}{c}{\textit{Plug and Play with Qwen-Image-Edit}} \\
        \midrule
        Qwen3-VL-8B~\cite{bai2025qwen2}  & 6.066 & 6.168 & 4.894 & 8.248 & 0.574 & 0.339 & 5.078 & 6.242\\
        Qwen3-VL-30B~\cite{bai2025qwen2}  & 4.092 & 4.284 & 2.904 & 8.270 & 0.558 & 0.296 & \underline{5.525} & 6.537\\
        GPT-5~\cite{openai2025gpt_4o}  & \textbf{7.498} & \textbf{7.560} & 6.060 & \textbf{8.616} & \underline{0.603} & 0.337 & 5.310 & \underline{6.546}\\
        \rowcolor{violet!6}
        \textbf{\NAME  3B} & 6.622 & 7.036 & \underline{6.244} & 8.206 & 0.592 & \underline{0.339} & 5.461 & 6.425\\
        \rowcolor{violet!10}
        \textbf{\NAME  7B} & \underline{6.882} & \underline{7.104} & \textbf{6.280}  & \underline{8.583} & \textbf{0.612} & \textbf{0.357} & \textbf{5.801} & \textbf{6.871}\\
        \bottomrule
    \end{tabular}
    }
    \vspace{-4mm}
\end{table*}
\textbf{Implement Details.}
All image-editing models are evaluated using their official pretrained checkpoints, producing 1024$\times$1024 outputs with default settings from their repositories to ensure a fair comparison.
\NAME acts as a lightweight plug-and-play reasoning layer that decomposes complex instructions and iteratively guides the base image-editing model toward the final result.

\noindent \textbf{Baseline Models.}
To comprehensively evaluate the effectiveness of \NAME, we conduct experiments on a diverse set of open-source and proprietary image editing models. For open-source baselines, we include InstructPix2Pix \cite{brooks2023instructpix2pix}, OmniGen2 \cite{wu2025omnigen2}, Bagel \cite{deng2025emerging}, Step1X-Edit \cite{liu2025step1x}, Flux.1-Kontext \cite{batifol2025flux}, and Qwen-Image-Edit \cite{wu2025qwen}.
For proprietary systems, we evaluate Seedream 4.0 \cite{seedream2025seedream}, GPT-Image \cite{openai2025gpt_image}, and Nano-Banana \cite{google2025nano_banana}. To assess \NAME’s enhancement capability, we integrate it with representative open-source instruction-guided image editing models, Step1X-Edit, Flux.1-Kontext-Dev, and Qwen-Image-Edit, forming \NAME-enhanced variants. Furthermore, we compare \NAME-enhanced models against other VLM-enhanced models, such as GPT-5~\cite{openai2025gpt_4o}, Qwen3-VL-8B-Instruct~\cite{bai2025qwen2}, and Qwen3-VL-30B-A3B-Instruct~\cite{bai2025qwen2}, under plug-and-play configurations with Flux.1-Kontext-Dev and Qwen-Image-Edit as base instruction-guided image editing models.

\noindent \textbf{Evaluation Benchmark.}
All models are evaluated on a challenging instruction-guided image editing benchmark built from the multi-turn subset of MagicBrush~\cite{zhang2023magicbrush} (following Section~\ref{sec:dataset}) and a subset of CompBench~\cite{jia2025compbench}, resulting in 500 test samples. Each sample includes rich, multi-sentence instructions requiring precise semantic alignment and high visual fidelity. For fair comparison, all evaluation images are standardized to 1024$\times$1024 resolution, and instruction lengths are limited to 77 words.

\noindent \textbf{Evaluation Metrics.}
We adopt two major evaluation dimensions: semantic consistency and perceptual quality. Semantic consistency is measured by GPT-SC, Gemini-SC, and Qwen3VL-SC, representing scores judged by GPT-5~\cite{openai2025gpt_4o}, Gemini-2.5-Flash~\cite{google2025gemini}, and Qwen3-VL-30B-A3B-Instruct~\cite{bai2025qwen2} respectively, following the viescore protocol \cite{ku2023viescore}. In addition, EditScore-SC provides an open-source semantic consistency metric \cite{luo2025editscore}. Perceptual quality is evaluated through ARNIQA~\cite{agnolucci2024arniqa} and TOPIQ~\cite{chen2024topiq}, while EditScore-PQ and EditScore-OA evaluate open-source perceptual and overall edit alignment quality~\cite{luo2025editscore}.
We evaluate models across two key dimensions: semantic consistency and perceptual quality.
\subsection{Evaluation of MIRA Performance}
\begin{table*}[t]
    \centering
    \caption{Comparison between MIRA with only SFT and \NAME further trained with GRPO after SFT. The additional GRPO stage consistently improves semantic consistency and perceptual quality across all backbones, with relative gains shown.}
    \label{tab:quan_compare_rl}
    \resizebox{\textwidth}{!}{
    \begin{tabular}{l|llllllll}
        \toprule
        \textbf{Method} & \makecell[c]{\textbf{GPT-SC}$\uparrow$}  & \makecell[c]{\textbf{Gemini-SC} $\uparrow$} & \makecell[c]{\textbf{Qwen3VL-SC} $\uparrow$} & \makecell[c]{\textbf{EditScore-SC} $\uparrow$} & \makecell[c]{\textbf{ARNIQA} $\uparrow$} & \makecell[c]{\textbf{TOPIQ} $\uparrow$}  & \makecell[c]{\textbf{EditScore-PQ} $\uparrow$}  & \makecell[c]{\textbf{EditScore-OA} $\uparrow$} \\
        \cmidrule(r){1-1} \cmidrule(r){2-5} \cmidrule(r){6-8} \cmidrule(r){9-9}
        Step1X-Edit + \NAME -SFT & 5.756 & 6.666 & 5.906 & 8.494 & 0.601 & 0.296 & 5.306 & 6.522\\
        \rowcolor{violet!10} Step1X-Edit + \NAME -GRPO & 6.322\textcolor{my_green}{$_{+9.83\%}$} & 6.886\textcolor{my_green}{$_{+3.30\%}$} & 5.844\textcolor{my_red}{$_{-1.05\%}$} & 8.668\textcolor{my_green}{$_{+2.05\%}$} & 0.610\textcolor{my_green}{$_{+1.50\%}$} & 0.304\textcolor{my_green}{$_{+2.70\%}$} & 5.640\textcolor{my_green}{$_{+6.29\%}$} & 6.800\textcolor{my_green}{$_{+4.26\%}$}\\
        Flux.1-Kontext + \NAME -SFT & 5.474 & 6.358 & 5.846 & 8.257 & 0.619 & 0.345 & 5.045 & 6.219\\
        \rowcolor{violet!10} Flux.1-Kontext + \NAME -GRPO & 6.202\textcolor{my_green}{$_{+13.30\%}$} & 6.670\textcolor{my_green}{$_{+4.91\%}$} & 5.802\textcolor{my_red}{$_{-0.75\%}$} & 8.532\textcolor{my_green}{$_{+3.33\%}$} & 0.619\textcolor{my_green}{$_{+0.00\%}$} & 0.353\textcolor{my_green}{$_{+2.32\%}$} & 5.473\textcolor{my_green}{$_{+8.48\%}$} & 6.610\textcolor{my_green}{$_{+6.29\%}$}\\
        Qwen-Image-Edit + \NAME -SFT & 5.992 & 6.730 & 6.196 & 8.491 & 0.605 & 0.349 & 5.625 & 6.703\\
        \rowcolor{violet!10} Qwen-Image-Edit + \NAME -GRPO & 6.882\textcolor{my_green}{$_{+14.85\%}$} & 7.104\textcolor{my_green}{$_{+5.56\%}$} & 6.244\textcolor{my_green}{$_{+0.77\%}$} & 8.583\textcolor{my_green}{$_{+1.08\%}$} & 0.612\textcolor{my_green}{$_{+1.16\%}$} & 0.357\textcolor{my_green}{$_{+2.29\%}$} & 5.801\textcolor{my_green}{$_{+3.13\%}$} & 6.871\textcolor{my_green}{$_{+2.51\%}$}\\
        \bottomrule
    \end{tabular}
    }

\end{table*}
\begin{figure*}[h]
  \centering
   \includegraphics[width=\textwidth]{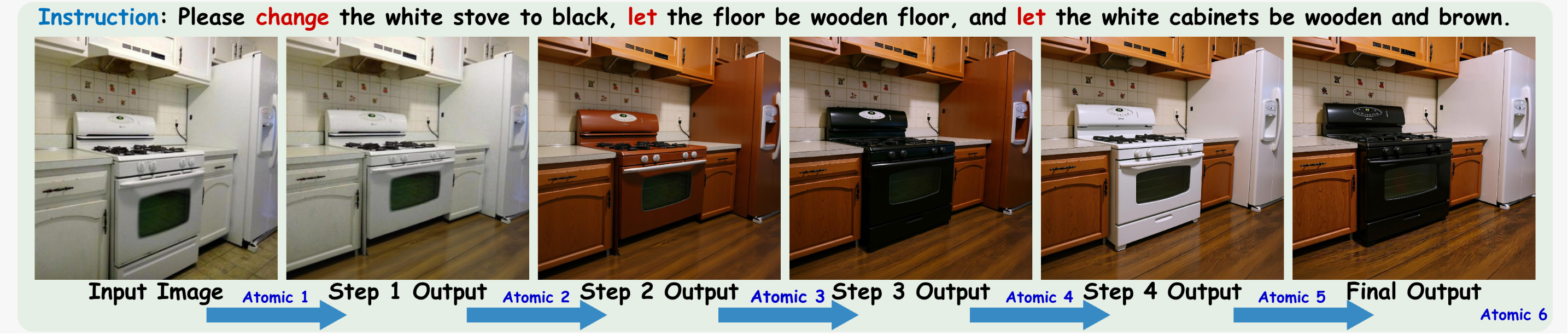}

   \caption{\textbf{Qualitative Case Study for \NAME's Error Mitigation Capability.} Atomic 1: \textit{Replace the floor to wooden floor.}, Atomic 2: \textit{Change the color of the white cabinet to wooden brown.}, Atomic 3: \textit{Change the color of the wooden stove to black.}, Atomic 4: \textit{Change the color of the wooden refrigerator to white.}, Atomic 5: \textit{Change the color of the white stove to black}, Atomic 6: \textit{\textlangle{}Stop\textrangle{}}.}
   \label{fig:mitigation}
\end{figure*}

\subsubsection{Quantitative Results}
Table \ref{tab:quan_compare_insert} reports results for proprietary editors, open-source instruction-guided editors, and their \NAME-enhanced variants. Across all open-source models, integrating \NAME consistently improves both semantic consistency and perceptual quality. For example, with Flux.1-Kontext, \NAME delivers gains of $4.98\%$, $6.04\%$, $13.41\%$, and $4.12\%$ on GPT-SC, Gemini-SC, Qwen3VL-SC, and EditScore-SC, respectively. Similar improvements appear for Step1X-Edit and Qwen-Image-Edit, where GPT-SC increases by $6.40\%$ and $9.41\%$. These results show that \NAME effectively enhances complex instruction following, enabling open-source editors to match or surpass proprietary systems such as GPT-Image and Nano-Banana.
Besides semantic gains, \NAME also improves perceptual quality, raising both ARNIQA and TOPIQ scores. Although multi-step editing might intuitively introduce diffusion noise or structural drift, we find the opposite: \NAME provides clearer, model-friendly instructions, reducing hallucination and artifact accumulation. Better-structured guidance allows the underlying generative priors to operate more stably, yielding edits that are both semantically faithful and visually cleaner.

Table~\ref{tab:quan_compare_vlm} further compares the same plug-and-play system equipped with different vision-language modules under identical settings, using Flux.1-Kontext-Dev and Qwen-Image-Edit as interchangeable external tools. Among Qwen3-VL-8B-Instruct, Qwen3-VL-30B-A3B-Instruct, GPT-5, \NAME 3B and \NAME 7B, \NAME 7B reaches the best balance of semantic consistency and perceptual quality across both editing models.
It outperforms open-source VLMs on almost every semantic metrics, with relative gains of $13\%$–$68\%$ on GPT-SC and $15\%$–$70\%$ on Gemini-SC, while achieving comparable or higher overall performance than GPT-5. 
This indicates that \NAME’s strength lies in multimodal reasoning for enhancing image editing, while its base model, Qwen2.5-VL-7B-Instruct is less capable in general reasoning and visual agentic capability than the Qwen3-VL series or GPT-5; its structured and iterative reasoning enables superior performance in instruction-guided image editing.

\subsubsection{Single-Turn Instruction Refinement}
Although \NAME is designed for multi-step iterative reasoning and editing, it also generalizes surprisingly well to single-turn instruction refinement, which is a task outside its training domain. When given an ambiguous instruction, \NAME can clarify the intent and rewrite it into a concise, executable editing instruction, enabling image editing models to follow user intent more reliably. 
We also provide a detailed analysis of how one-shot refinement leads to consistent gains across diverse editing categories in the supplementary material.

\subsection{Ablation Study}
\subsubsection{Effectiveness of RL}

To quantify the impact of reinforcement learning on multimodal reasoning and editing quality, we conduct an ablation comparing \NAME trained only with supervised fine-tuning  against the full two-stage version incorporating Group Relative Policy Optimization, as illustrated in Figure \ref{fig:training_pipeline}. In this setting, \NAME-SFT corresponds to the policy learned purely by imitation from curated instruction–image pairs, while \NAME-GRPO further refines this policy through reward-driven optimization, using customized composite rewards derived from an image editing model and a reward model for image editing. As shown in Table \ref{tab:quan_compare_rl}, RL post-training with our customized reward function yields consistent performance gains across nearly all metrics and open-source image editing models listed. For example, Step1X-Edit + \NAME-GRPO achieves $9.83 \%$, $6.29\%$, and $4.26\%$ improvements on GPT-SC, EditScore-PQ, and EditScore-OA, respectively, compared to its SFT counterpart. Similar trends are observed for plugging-and-playing with Flux.1-Kontext and Qwen-Image-Edit, indicating that the post-training stage effectively enhances both semantic consistency and perceptual quality. These results validate that reward-driven optimization enables \NAME to learn more fine-grained multimodal reasoning policies, producing editing instructions that are semantically faithful, visually coherent, better aligned with human-preferred editing outcomes, and easier for image editing models to understand.

\subsubsection{Reasoning Steps and Excution Quality}

To evaluate whether a larger reasoning budget improves execution quality, we vary the maximum number of inference steps from 3 to 7 and report the corresponding performance in Table~\ref{tab:max_step_ablation}. Across both semantic consistency and perceptual quality metrics, we observe only minor variations as the maximum step budget increases. Allowing additional steps yields modest improvements, most notably in semantic consistency, suggesting that \NAME occasionally benefits from the extra reasoning capacity when handling more challenging edits. However, the overall changes remain small, indicating that iterative reasoning depth is not the primary factor governing final output quality.

\begin{table*}[t]
    \centering
    \caption{\textbf{Ablation Study on the Impact of Maximum Inference Steps on \NAME’s Editing Performance.} We evaluate Flux.1-Kontext + \NAME under different maximum number of inference step (from 3 to 7) and report semantic consistency and perceptual quality across multiple metrics. The \textbf{best} and \underline{second best} results are in bold and underlined, respectively.}
    \label{tab:max_step_ablation}
    \resizebox{\textwidth}{!}{
    \begin{tabular}{l|cccccccc}
        \toprule
        \textbf{Method} & \textbf{GPT-SC} $\uparrow$ & \textbf{Gemini-SC} $\uparrow$ &
        \textbf{Qwen3VL-SC} $\uparrow$ & \textbf{EditScore-SC} $\uparrow$ &
        \textbf{ARNIQA} $\uparrow$ & \textbf{TOPIQ} $\uparrow$ &
        \textbf{EditScore-PQ} $\uparrow$ & \textbf{EditScore-OA} $\uparrow$ \\
        \cmidrule(r){1-1} \cmidrule(r){2-5} \cmidrule(r){6-8} \cmidrule(r){9-9}
        
        \rowcolor{violet!6}
        Flux.1-Kontext + \NAME + Max=3 & 5.912 & 6.626 & 5.788 & \textbf{8.659} & 0.611 & 0.346 & \textbf{5.550} & \textbf{6.785}\\

        \rowcolor{violet!8}
        Flux.1-Kontext + \NAME + Max=4 & 6.204 & 6.284 & 5.800 & \underline{8.553} & 0.614 & 0.350 & 5.410 & \underline{6.654} \\

        \rowcolor{violet!10}
        Flux.1-Kontext + \NAME + Max=5 & 6.202 & 6.670 & 5.802 & 8.532 & \textbf{0.619} & \textbf{0.353} & \underline{5.473} & 6.610 \\

        \rowcolor{violet!12}
        Flux.1-Kontext + \NAME + Max=6 & \underline{6.304} & \textbf{6.760} & \underline{5.840} & 8.487 & 0.617 & \underline{0.352} & 5.406 & 6.628 \\

        \rowcolor{violet!14}
        Flux.1-Kontext + \NAME + Max=7 & \textbf{6.320} & \underline{6.730} & \textbf{5.892} & 8.480 & \underline{0.618} & \textbf{0.353} & 5.337 & 6.556 \\
        
        \bottomrule
    \end{tabular}
    }
    \vspace{-3mm}
\end{table*}

\subsection{Latency and Computational Cost Analysis}
Figure~\ref{fig:qual} and Table~\ref{tab:quan_compare_insert} illustrate that \NAME’s iterative reasoning markedly boosts editing quality, though at the expense of additional latency and computation. We analyze these overheads in this section.

All latency measurements were conducted on a single NVIDIA H100 GPU. For a fair comparison, we report the time required to generate a single 1024x1024 edited image. Using MIRA 7B paired with Flux.1-Kontext, and averaged across 500 benchmark samples, each MIRA reasoning step takes 0.746 seconds, while a single Flux.1-Kontext inference takes 14.445 seconds. The framework performs an average of 4.111 iterations before issuing a stop signal, yielding a total end-to-end latency of $48.005$ seconds. To contextualize this cost, we compare against two representative proprietary systems: GPT-Image requires 71.7 seconds and \$0.17 per edit, while Nano-Banana achieves 12.3 seconds at \$0.04 per edit. Although our iterative framework introduces additional latency, it remains practical, especially given that MIRA is fully open-source, and it provides substantial gains in semantic consistency and visual fidelity.
\subsection{Inherent Robustness and Error Mitigation}
A key strength of \NAME's iterative framework is its inherent robustness against error accumulation. Unlike other agentic frameworks that generate a static and open-loop plan, \NAME performs stateful and closed-loop reasoning. At each step, it makes its decision by re-evaluating the latest editing output $I_{t-1}$ against the constant goal defined by the input image $I_0$ and the constant goal defined by $C$. This allows MIRA to implicitly mitigate or correct minor deviations or execution errors from previous steps. Figure~\ref{fig:mitigation} provides a representative example illustrating \NAME's inherent robustness to execution errors made by the external instruction-guided image editing model. Given the complex instruction: ``Change the white stove to black, let the floor be wooden, and let the white cabinets be wooden and brown'', \NAME decomposes the request into a sequence of visually grounded atomic edits and executes iteratively.

In the first step (Atomic 1), \NAME successfully guides the base editing model to convert the floor into a wooden texture. However, during Atomic 2, although all the white cabinets are correctly recolored to a wooden and brown appearance, the refrigerator, whose color should remain white, is mistakenly changed to brown. It means that the editing model produces an incorrect or partially incorrect transformation. 
Because \NAME operates in a closed-loop manner, it does not assume the correctness of any intermediate step. Instead, at each iteration, it re-analyzes the current intermediate image $I_{t-1}$ together with the original image $I_0$ and the input complex instruction $C$. 
After executing Atomic 3, \NAME detects that the refrigerator still does not match the intended appearance and therefore issues a corrective atomic instruction in Atomic 4 (Change the wooden refrigerator to white). However, upon inspecting the result of Step 4, \NAME identifies a new inconsistency: the stove, which should remain black, has been unintentionally changed to white during the previous correction. Consequently, \NAME generates another corrective instruction in Atomic 5 (Change the white stove to black).

This demo instruction reflects an intentional error-mitigation strategy emerging from \NAME’s state-conditioned reasoning: the agent continuously verifies the current state and issues targeted corrections whenever deviations appear. After applying this final corrective action and confirming that the output image fully satisfies the instruction, \NAME emits the \texttt{\textlangle Stop\textrangle} signal in Atomic 6 to terminate the editing process.

These results indicate that \NAME can effectively diagnose discrepancies introduced by editing models and dynamically adjust their editing trajectory, mitigating error propagation and preserving alignment with user intent.
\subsection{Reliability of the Termination Mechanism}

Table~\ref{tab:termination} reports the average number of actual reasoning steps taken by \NAME under different maximum step budgets. As the allowed iterations increase from 3 to 5, the average step count rises moderately (from 2.976 to 4.111). This trend reflects the fact that many instructions in our benchmark consist of multiple sub-tasks, often requiring at least three editing actions to complete. However, when the maximum step budget is further expanded from 5 to 7, the average actual steps remain nearly unchanged (4.096 to 4.208), indicating no meaningful scaling with the larger budget.
These results show that \NAME’s termination behavior is goal-driven rather than budget-driven. Specifically:
\begin{itemize}
\item It refrains from using the full step budget, even when additional iterations are available.
\item It mitigates over-editing while improveing the reliability.
\item It consistently converges to a stable number of reasoning steps on this benchmark (
$\sim$ 4 steps on average).
\end{itemize}

\begin{table}[t]
    \centering
    \caption{\textbf{Ablation Study on the Reliability of \NAME’s Termination Mechanism.} We report the average number of inference steps taken by \NAME across our 500-sample benchmark when paired with Flux.1-Kontext under different maximum allowed inference step budgets.}
    \vspace{-2mm}
    \label{tab:termination}
    \resizebox{\linewidth}{!}{
    \begin{tabular}{l|ccccc}
    \toprule
    \textbf{Method} & \textbf{Max=3} & \textbf{Max=4} & \textbf{Max=5} & \textbf{Max=6} & \textbf{Max=7} \\
    \cmidrule(r){1-1} \cmidrule(r){2-6}
    Flux.1-Kontext + MIRA & 2.976 & 3.672 & 4.111 & 4.096 & 4.208 \\
    \bottomrule
    \end{tabular}
    }
    \vspace{-4mm}
\end{table}

%% file: sec/6_conclusion.tex
\section{Conclusion}
\label{sec:conclusion}

We introduce \NAME, a lightweight multimodal reasoning agent that reframes image editing as an iterative perception–reasoning–action process. By generating atomic edit instructions and adapting with stepwise feedback, \NAME greatly enhances the capability of open-source image editing models on complex instruction-guided editing tasks. In addition, we also propose a novel large-scale dataset, \DATA, for training automatic tool-use models for instruction-guided image editing. Supported by \DATA and GRPO-based optimization with a customized reward function, \NAME consistently improves semantic consistency and perceptual quality, narrowing the gap with proprietary systems. Empirical results highlight iterative multimodal reasoning as an effective, scalable, and novel paradigm for controllable and high-quality image editing.

\section*{Acknowledgements}
This work was supported by Goergen Institute for Data Science and Artificial Intelligence at University of Rochester.

%% file: main.bib
@String(ICME = {Int. Conf. Multimedia and Expo})

@String(AAAI = {AAAI})

@String(ICME  =	{ICME})

@article{hertz2022prompt,
  title={Prompt-to-prompt image editing with cross attention control},
  author={Hertz, Amir and Mokady, Ron and Tenenbaum, Jay and Aberman, Kfir and Pritch, Yael and Cohen-Or, Daniel},
  journal={arXiv preprint arXiv:2208.01626},
  year={2022}
}

@article{meng2021sdedit,
  title={Sdedit: Guided image synthesis and editing with stochastic differential equations},
  author={Meng, Chenlin and He, Yutong and Song, Yang and Song, Jiaming and Wu, Jiajun and Zhu, Jun-Yan and Ermon, Stefano},
  journal={arXiv preprint arXiv:2108.01073},
  year={2021}
}

@article{couairon2022diffedit,
  title={Diffedit: Diffusion-based semantic image editing with mask guidance},
  author={Couairon, Guillaume and Verbeek, Jakob and Schwenk, Holger and Cord, Matthieu},
  journal={arXiv preprint arXiv:2210.11427},
  year={2022}
}

@inproceedings{kawar2023imagic,
  title={Imagic: Text-based real image editing with diffusion models},
  author={Kawar, Bahjat and Zada, Shiran and Lang, Oran and Tov, Omer and Chang, Huiwen and Dekel, Tali and Mosseri, Inbar and Irani, Michal},
  booktitle={Proceedings of the IEEE/CVF conference on computer vision and pattern recognition},
  pages={6007--6017},
  year={2023}
}

@inproceedings{brooks2023instructpix2pix,
  title={Instructpix2pix: Learning to follow image editing instructions},
  author={Brooks, Tim and Holynski, Aleksander and Efros, Alexei A},
  booktitle={Proceedings of the IEEE/CVF conference on computer vision and pattern recognition},
  pages={18392--18402},
  year={2023}
}

@article{gan2023instructcv,
  title={Instructcv: Instruction-tuned text-to-image diffusion models as vision generalists},
  author={Gan, Yulu and Park, Sungwoo and Schubert, Alexander and Philippakis, Anthony and Alaa, Ahmed M},
  journal={arXiv preprint arXiv:2310.00390},
  year={2023}
}

@article{yu2024promptfix,
  title={Promptfix: You prompt and we fix the photo},
  author={Yu, Yongsheng and Zeng, Ziyun and Hua, Hang and Fu, Jianlong and Luo, Jiebo},
  journal={arXiv preprint arXiv:2405.16785},
  year={2024}
}

@article{zhang2023magicbrush,
  title={Magicbrush: A manually annotated dataset for instruction-guided image editing},
  author={Zhang, Kai and Mo, Lingbo and Chen, Wenhu and Sun, Huan and Su, Yu},
  journal={Advances in Neural Information Processing Systems},
  volume={36},
  pages={31428--31449},
  year={2023}
}

@article{zhao2024ultraedit,
  title={Ultraedit: Instruction-based fine-grained image editing at scale},
  author={Zhao, Haozhe and Ma, Xiaojian Shawn and Chen, Liang and Si, Shuzheng and Wu, Rujie and An, Kaikai and Yu, Peiyu and Zhang, Minjia and Li, Qing and Chang, Baobao},
  journal={Advances in Neural Information Processing Systems},
  volume={37},
  pages={3058--3093},
  year={2024}
}

@inproceedings{geng2024instructdiffusion,
  title={Instructdiffusion: A generalist modeling interface for vision tasks},
  author={Geng, Zigang and Yang, Binxin and Hang, Tiankai and Li, Chen and Gu, Shuyang and Zhang, Ting and Bao, Jianmin and Zhang, Zheng and Li, Houqiang and Hu, Han and others},
  booktitle={Proceedings of the IEEE/CVF Conference on computer vision and pattern recognition},
  pages={12709--12720},
  year={2024}
}

@article{wang2023instructedit,
  title={Instructedit: Improving automatic masks for diffusion-based image editing with user instructions},
  author={Wang, Qian and Zhang, Biao and Birsak, Michael and Wonka, Peter},
  journal={arXiv preprint arXiv:2305.18047},
  year={2023}
}

@article{fu2023guiding,
  title={Guiding instruction-based image editing via multimodal large language models},
  author={Fu, Tsu-Jui and Hu, Wenze and Du, Xianzhi and Wang, William Yang and Yang, Yinfei and Gan, Zhe},
  journal={arXiv preprint arXiv:2309.17102},
  year={2023}
}

@inproceedings{huang2024smartedit,
  title={Smartedit: Exploring complex instruction-based image editing with multimodal large language models},
  author={Huang, Yuzhou and Xie, Liangbin and Wang, Xintao and Yuan, Ziyang and Cun, Xiaodong and Ge, Yixiao and Zhou, Jiantao and Dong, Chao and Huang, Rui and Zhang, Ruimao and others},
  booktitle={Proceedings of the IEEE/CVF Conference on Computer Vision and Pattern Recognition},
  pages={8362--8371},
  year={2024}
}

@article{wu2025omnigen2,
  title={OmniGen2: Exploration to Advanced Multimodal Generation},
  author={Wu, Chenyuan and Zheng, Pengfei and Yan, Ruiran and Xiao, Shitao and Luo, Xin and Wang, Yueze and Li, Wanli and Jiang, Xiyan and Liu, Yexin and Zhou, Junjie and others},
  journal={arXiv preprint arXiv:2506.18871},
  year={2025}
}

@article{wu2025qwen,
  title={Qwen-image technical report},
  author={Wu, Chenfei and Li, Jiahao and Zhou, Jingren and Lin, Junyang and Gao, Kaiyuan and Yan, Kun and Yin, Sheng-ming and Bai, Shuai and Xu, Xiao and Chen, Yilei and others},
  journal={arXiv preprint arXiv:2508.02324},
  year={2025}
}

@article{liu2025step1x,
  title={Step1x-edit: A practical framework for general image editing},
  author={Liu, Shiyu and Han, Yucheng and Xing, Peng and Yin, Fukun and Wang, Rui and Cheng, Wei and Liao, Jiaqi and Wang, Yingming and Fu, Honghao and Han, Chunrui and others},
  journal={arXiv preprint arXiv:2504.17761},
  year={2025}
}

@article{batifol2025flux,
  title={FLUX. 1 Kontext: Flow Matching for In-Context Image Generation and Editing in Latent Space},
  author={Batifol, Stephen and Blattmann, Andreas and Boesel, Frederic and Consul, Saksham and Diagne, Cyril and Dockhorn, Tim and English, Jack and English, Zion and Esser, Patrick and Kulal, Sumith and others},
  journal={arXiv e-prints},
  pages={arXiv--2506},
  year={2025}
}

@misc{google2025nano_banana,
  title        = {Nano Banana},
  author       = {Google AI}}

@misc{google2025gemini,
  title        = {Gemini 2.5},
  author       = {Google AI}}

@misc{openai2025gpt_image,
  title        = {GPT-Image},
  author       = {OpenAI},
  year         = {2025},
  howpublished = {\url{https://openai.com}},  
}

@misc{openai2025gpt_4o,
  title        = {ChatGPT},
  author       = {OpenAI},
  year         = {2024},
  howpublished = {\url{https://openai.com}},  
}

@article{seedream2025seedream,
  title={Seedream 4.0: Toward next-generation multimodal image generation},
  author={Seedream, Team and Chen, Yunpeng and Gao, Yu and Gong, Lixue and Guo, Meng and Guo, Qiushan and Guo, Zhiyao and Hou, Xiaoxia and Huang, Weilin and Huang, Yixuan and others},
  journal={arXiv preprint arXiv:2509.20427},
  year={2025}
}

@article{huang2024dialoggen,
  title={Dialoggen: Multi-modal interactive dialogue system for multi-turn text-to-image generation},
  author={Huang, Minbin and Long, Yanxin and Deng, Xinchi and Chu, Ruihang and Xiong, Jiangfeng and Liang, Xiaodan and Cheng, Hong and Lu, Qinglin and Liu, Wei},
  journal={arXiv preprint arXiv:2403.08857},
  year={2024}
}

@article{wang2024genartist,
  title={Genartist: Multimodal llm as an agent for unified image generation and editing},
  author={Wang, Zhenyu and Li, Aoxue and Li, Zhenguo and Liu, Xihui},
  journal={Advances in Neural Information Processing Systems},
  volume={37},
  pages={128374--128395},
  year={2024}
}

@article{chen2025t2i,
  title={T2I-Copilot: A Training-Free Multi-Agent Text-to-Image System for Enhanced Prompt Interpretation and Interactive Generation},
  author={Chen, Chieh-Yun and Shi, Min and Zhang, Gong and Shi, Humphrey},
  journal={arXiv preprint arXiv:2507.20536},
  year={2025}
}

@article{cai2025hidream,
  title={HiDream-I1: A High-Efficient Image Generative Foundation Model with Sparse Diffusion Transformer},
  author={Cai, Qi and Chen, Jingwen and Chen, Yang and Li, Yehao and Long, Fuchen and Pan, Yingwei and Qiu, Zhaofan and Zhang, Yiheng and Gao, Fengbin and Xu, Peihan and others},
  journal={arXiv preprint arXiv:2505.22705},
  year={2025}
}

@article{sun2025latent,
  title={Latent Chain-of-Thought for Visual Reasoning},
  author={Sun, Guohao and Hua, Hang and Wang, Jian and Luo, Jiebo and Dianat, Sohail and Rabbani, Majid and Rao, Raghuveer and Tao, Zhiqiang},
  journal={arXiv preprint arXiv:2510.23925},
  year={2025}
}

@inproceedings{xia2024llmga,
  title={Llmga: Multimodal large language model based generation assistant},
  author={Xia, Bin and Wang, Shiyin and Tao, Yingfan and Wang, Yitong and Jia, Jiaya},
  booktitle={European Conference on Computer Vision},
  pages={389--406},
  year={2024},
  organization={Springer}
}

@inproceedings{hu2023promptcap,
  title={Promptcap: Prompt-guided image captioning for vqa with gpt-3},
  author={Hu, Yushi and Hua, Hang and Yang, Zhengyuan and Shi, Weijia and Smith, Noah A and Luo, Jiebo},
  booktitle={Proceedings of the IEEE/CVF International Conference on Computer Vision},
  pages={2963--2975},
  year={2023}
}

@article{zhu2025internvl3,
  title={Internvl3: Exploring advanced training and test-time recipes for open-source multimodal models},
  author={Zhu, Jinguo and Wang, Weiyun and Chen, Zhe and Liu, Zhaoyang and Ye, Shenglong and Gu, Lixin and Tian, Hao and Duan, Yuchen and Su, Weijie and Shao, Jie and others},
  journal={arXiv preprint arXiv:2504.10479},
  year={2025}
}

@article{meta2025llama,
  title={The llama 4 herd: The beginning of a new era of natively multimodal ai innovation},
  author={Meta, AI},
  journal={https://ai. meta. com/blog/llama-4-multimodal-intelligence/, checked on},
  volume={4},
  number={7},
  pages={2025},
  year={2025}
}

@inproceedings{hua2025finecaption,
  title={Finecaption: Compositional image captioning focusing on wherever you want at any granularity},
  author={Hua, Hang and Liu, Qing and Zhang, Lingzhi and Shi, Jing and Kim, Soo Ye and Zhang, Zhifei and Wang, Yilin and Zhang, Jianming and Lin, Zhe and Luo, Jiebo},
  booktitle={Proceedings of the Computer Vision and Pattern Recognition Conference},
  pages={24763--24773},
  year={2025}
}

@article{liang2025llm,
  title={An LLM-LVLM Driven Agent for Iterative and Fine-Grained Image Editing},
  author={Liang, Zihan and Sun, Jiahao and Ma, Haoran},
  journal={arXiv preprint arXiv:2508.17435},
  year={2025}
}

@inproceedings{hua2025v2xum,
  title={V2xum-llm: Cross-modal video summarization with temporal prompt instruction tuning},
  author={Hua, Hang and Tang, Yunlong and Xu, Chenliang and Luo, Jiebo},
  booktitle={Proceedings of the AAAI Conference on Artificial Intelligence},
  volume={39},
  number={4},
  pages={3599--3607},
  year={2025}
}

@article{yeh2025beyond,
  title={Beyond Simple Edits: X-Planner for Complex Instruction-Based Image Editing},
  author={Yeh, Chun-Hsiao and Wang, Yilin and Zhao, Nanxuan and Zhang, Richard and Li, Yuheng and Ma, Yi and Singh, Krishna Kumar},
  journal={arXiv preprint arXiv:2507.05259},
  year={2025}
}

@inproceedings{hua2024finematch,
  title={Finematch: Aspect-based fine-grained image and text mismatch detection and correction},
  author={Hua, Hang and Shi, Jing and Kafle, Kushal and Jenni, Simon and Zhang, Daoan and Collomosse, John and Cohen, Scott and Luo, Jiebo},
  booktitle={European Conference on Computer Vision},
  pages={474--491},
  year={2024},
  organization={Springer}
}

@article{gupta2025costa,
  title={CoSTA*: Cost-Sensitive Toolpath Agent for Multi-turn Image Editing},
  author={Gupta, Advait and Velaga, NandaKiran and Nguyen, Dang and Zhou, Tianyi},
  journal={arXiv preprint arXiv:2503.10613},
  year={2025}
}

@article{zhang2025context,
  title={In-context edit: Enabling instructional image editing with in-context generation in large scale diffusion transformer},
  author={Zhang, Zechuan and Xie, Ji and Lu, Yu and Yang, Zongxin and Yang, Yi},
  journal={arXiv preprint arXiv:2504.20690},
  year={2025}
}

@article{hu2025image,
  title={Image Editing As Programs with Diffusion Models},
  author={Hu, Yujia and Liu, Songhua and Tan, Zhenxiong and Yang, Xingyi and Wang, Xinchao},
  journal={arXiv preprint arXiv:2506.04158},
  year={2025}
}

@inproceedings{ji2025instruction,
  title={Instruction-based Image Editing with Planning, Reasoning, and Generation},
  author={Ji, Liya and Qi, Chenyang and Chen, Qifeng},
  booktitle={Proceedings of the IEEE/CVF International Conference on Computer Vision},
  pages={17506--17515},
  year={2025}
}

@article{deng2025emerging,
  title={Emerging properties in unified multimodal pretraining},
  author={Deng, Chaorui and Zhu, Deyao and Li, Kunchang and Gou, Chenhui and Li, Feng and Wang, Zeyu and Zhong, Shu and Yu, Weihao and Nie, Xiaonan and Song, Ziang and others},
  journal={arXiv preprint arXiv:2505.14683},
  year={2025}
}

@article{ku2023viescore,
  title={Viescore: Towards explainable metrics for conditional image synthesis evaluation},
  author={Ku, Max and Jiang, Dongfu and Wei, Cong and Yue, Xiang and Chen, Wenhu},
  journal={arXiv preprint arXiv:2312.14867},
  year={2023}
}

@article{luo2025editscore,
  title={EditScore: Unlocking Online RL for Image Editing via High-Fidelity Reward Modeling},
  author={Luo, Xin and Wang, Jiahao and Wu, Chenyuan and Xiao, Shitao and Jiang, Xiyan and Lian, Defu and Zhang, Jiajun and Liu, Dong and others},
  journal={arXiv preprint arXiv:2509.23909},
  year={2025}
}

@inproceedings{zhou2025fireedit,
  title={Fireedit: Fine-grained instruction-based image editing via region-aware vision language model},
  author={Zhou, Jun and Li, Jiahao and Xu, Zunnan and Li, Hanhui and Cheng, Yiji and Hong, Fa-Ting and Lin, Qin and Lu, Qinglin and Liang, Xiaodan},
  booktitle={Proceedings of the Computer Vision and Pattern Recognition Conference},
  pages={13093--13103},
  year={2025}
}

@article{bai2025qwen2,
  title={Qwen2. 5-vl technical report},
  author={Bai, Shuai and Chen, Keqin and Liu, Xuejing and Wang, Jialin and Ge, Wenbin and Song, Sibo and Dang, Kai and Wang, Peng and Wang, Shijie and Tang, Jun and others},
  journal={arXiv preprint arXiv:2502.13923},
  year={2025}
}

@article{jia2025compbench,
  title={CompBench: Benchmarking Complex Instruction-guided Image Editing},
  author={Jia, Bohan and Huang, Wenxuan and Tang, Yuntian and Qiao, Junbo and Liao, Jincheng and Cao, Shaosheng and Zhao, Fei and Feng, Zhaopeng and Gu, Zhouhong and Yin, Zhenfei and others},
  journal={arXiv preprint arXiv:2505.12200},
  year={2025}
}

@inproceedings{agnolucci2024arniqa,
  title={Arniqa: Learning distortion manifold for image quality assessment},
  author={Agnolucci, Lorenzo and Galteri, Leonardo and Bertini, Marco and Del Bimbo, Alberto},
  booktitle={Proceedings of the IEEE/CVF Winter Conference on Applications of Computer Vision},
  pages={189--198},
  year={2024}
}

@article{chen2024topiq,
  title={Topiq: A top-down approach from semantics to distortions for image quality assessment},
  author={Chen, Chaofeng and Mo, Jiadi and Hou, Jingwen and Wu, Haoning and Liao, Liang and Sun, Wenxiu and Yan, Qiong and Lin, Weisi},
  journal={IEEE Transactions on Image Processing},
  volume={33},
  pages={2404--2418},
  year={2024},
  publisher={IEEE}
}

@article{ge2024seed,
  title={Seed-data-edit technical report: A hybrid dataset for instructional image editing},
  author={Ge, Yuying and Zhao, Sijie and Li, Chen and Ge, Yixiao and Shan, Ying},
  journal={arXiv preprint arXiv:2405.04007},
  year={2024}
}

@article{shao2024deepseekmath,
  title={Deepseekmath: Pushing the limits of mathematical reasoning in open language models},
  author={Shao, Zhihong and Wang, Peiyi and Zhu, Qihao and Xu, Runxin and Song, Junxiao and Bi, Xiao and Zhang, Haowei and Zhang, Mingchuan and Li, YK and Wu, Yang and others},
  journal={arXiv preprint arXiv:2402.03300},
  year={2024}
}

@article{zeng2025glm,
  title={Glm-4.5: Agentic, reasoning, and coding (arc) foundation models},
  author={Zeng, Aohan and Lv, Xin and Zheng, Qinkai and Hou, Zhenyu and Chen, Bin and Xie, Chengxing and Wang, Cunxiang and Yin, Da and Zeng, Hao and Zhang, Jiajie and others},
  journal={arXiv preprint arXiv:2508.06471},
  year={2025}
}

@article{guo2025deepseek,
  title={Deepseek-r1: Incentivizing reasoning capability in llms via reinforcement learning},
  author={Guo, Daya and Yang, Dejian and Zhang, Haowei and Song, Junxiao and Zhang, Ruoyu and Xu, Runxin and Zhu, Qihao and Ma, Shirong and Wang, Peiyi and Bi, Xiao and others},
  journal={arXiv preprint arXiv:2501.12948},
  year={2025}
}

@article{lin2024battleagent,
  title={Battleagent: Multi-modal dynamic emulation on historical battles to complement historical analysis},
  author={Lin, Shuhang and Hua, Wenyue and Li, Lingyao and Chang, Che-Jui and Fan, Lizhou and Ji, Jianchao and Hua, Hang and Jin, Mingyu and Luo, Jiebo and Zhang, Yongfeng},
  journal={arXiv preprint arXiv:2404.15532},
  year={2024}
}

@article{cao2024presto,
  title={PRESTO: Progressive pretraining enhances synthetic chemistry outcomes},
  author={Cao, He and Shao, Yanjun and Liu, Zhiyuan and Liu, Zijing and Tang, Xiangru and Yao, Yuan and Li, Yu},
  journal={arXiv preprint arXiv:2406.13193},
  year={2024}
}

@article{tang2025chemagent,
  title={Chemagent: Self-updating library in large language models improves chemical reasoning},
  author={Tang, Xiangru and Hu, Tianyu and Ye, Muyang and Shao, Yanjun and Yin, Xunjian and Ouyang, Siru and Zhou, Wangchunshu and Lu, Pan and Zhang, Zhuosheng and Zhao, Yilun and others},
  journal={arXiv preprint arXiv:2501.06590},
  year={2025}
}

@article{tang2025medagentsbench,
  title={Medagentsbench: Benchmarking thinking models and agent frameworks for complex medical reasoning},
  author={Tang, Xiangru and Shao, Daniel and Sohn, Jiwoong and Chen, Jiapeng and Zhang, Jiayi and Xiang, Jinyu and Wu, Fang and Zhao, Yilun and Wu, Chenglin and Shi, Wenqi and others},
  journal={arXiv preprint arXiv:2503.07459},
  year={2025}
}

@article{tang2025cellforge,
  title={Cellforge: Agentic design of virtual cell models},
  author={Tang, Xiangru and Yu, Zhuoyun and Chen, Jiapeng and Cui, Yan and Shao, Daniel and Wang, Weixu and Wu, Fang and Zhuang, Yuchen and Shi, Wenqi and Huang, Zhi and others},
  journal={arXiv preprint arXiv:2508.02276},
  year={2025}
}

@article{yu2025omnipaint,
  title={Omnipaint: Mastering object-oriented editing via disentangled insertion-removal inpainting},
  author={Yu, Yongsheng and Zeng, Ziyun and Zheng, Haitian and Luo, Jiebo},
  journal={arXiv preprint arXiv:2503.08677},
  year={2025}
}

@inproceedings{song2023objectstitch,
  title={Objectstitch: Object compositing with diffusion model},
  author={Song, Yizhi and Zhang, Zhifei and Lin, Zhe and Cohen, Scott and Price, Brian and Zhang, Jianming and Kim, Soo Ye and Aliaga, Daniel},
  booktitle={Proceedings of the IEEE/CVF Conference on Computer Vision and Pattern Recognition},
  pages={18310--18319},
  year={2023}
}

@inproceedings{song2024imprint,
  title={Imprint: Generative object compositing by learning identity-preserving representation},
  author={Song, Yizhi and Zhang, Zhifei and Lin, Zhe and Cohen, Scott and Price, Brian and Zhang, Jianming and Kim, Soo Ye and Zhang, He and Xiong, Wei and Aliaga, Daniel},
  booktitle={Proceedings of the IEEE/CVF Conference on Computer Vision and Pattern Recognition},
  pages={8048--8058},
  year={2024}
}

@article{song2024refine,
  title={Refine-by-align: Reference-guided artifacts refinement through semantic alignment},
  author={Song, Yizhi and He, Liu and Zhang, Zhifei and Kim, Soo Ye and Zhang, He and Xiong, Wei and Lin, Zhe and Price, Brian and Cohen, Scott and Zhang, Jianming and others},
  journal={arXiv preprint arXiv:2412.00306},
  year={2024}
}

@inproceedings{zhengpixperfect,
  title={PixPerfect: Seamless Latent Diffusion Local Editing with Discriminative Pixel-Space Refinement},
  author={Zheng, Haitian and Yao, Yuan and Yu, Yongsheng and Zhou, Yuqian and Luo, Jiebo and Lin, Zhe},
  booktitle={The Thirty-ninth Annual Conference on Neural Information Processing Systems},
  year={2025}
}

@article{hua2024mmcomposition,
  title={Mmcomposition: Revisiting the compositionality of pre-trained vision-language models},
  author={Hua, Hang and Tang, Yunlong and Zeng, Ziyun and Cao, Liangliang and Yang, Zhengyuan and He, Hangfeng and Xu, Chenliang and Luo, Jiebo},
  journal={arXiv preprint arXiv:2410.09733},
  year={2024}
}

@article{hua2025mmig,
  title={MMIG-Bench: Towards Comprehensive and Explainable Evaluation of Multi-Modal Image Generation Models},
  author={Hua, Hang and Zeng, Ziyun and Song, Yizhi and Tang, Yunlong and He, Liu and Aliaga, Daniel and Xiong, Wei and Luo, Jiebo},
  journal={arXiv preprint arXiv:2505.19415},
  year={2025}
}

@inproceedings{yu2024chain,
  title={Chain-of-thought prompting for demographic inference with large multimodal models},
  author={Yu, Yongsheng and Luo, Jiebo},
  booktitle={2024 IEEE International Conference on Multimedia and Expo (ICME)},
  pages={1--7},
  year={2024},
  organization={IEEE}
}

@article{tang2025video,
  title={Video-lmm post-training: A deep dive into video reasoning with large multimodal models},
  author={Tang, Yunlong and Bi, Jing and Liu, Pinxin and Pan, Zhenyu and Tan, Zhangyun and Shen, Qianxiang and Liu, Jiani and Hua, Hang and Guo, Junjia and Xiao, Yunzhong and others},
  journal={arXiv preprint arXiv:2510.05034},
  year={2025}
}

@article{zeng2026automated,
  title={Automated Detection and Quantitative Assessment of Dental Plaque in Intraoral Images},
  author={Zeng, Ziyun and Chen, Junyu and Rashwan, Noha and Al Jallad, Nisreen and Xiao, Jin and Luo, Jiebo},
  journal={ACM Transactions on Computing for Healthcare},
  year={2026},
  publisher={ACM New York, NY}
}

@article{zeng2025use,
  title={Use of artificial intelligence to detect dental caries on intraoral photos},
  author={Zeng, Ziyun and Ramesh, Ashwin and Ruan, Jinglong and Hao, Peirong and Al\_Jallad, N and Jang, Hoonji and Ly-Mapes, Oriana and Fiscella, Kevin and Xiao, Jin and Luo, Jiebo},
  journal={Quintessence international},
  year={2025},
  publisher={Quintessence Publisher}
}
